\documentclass[runningheads]{llncs}

\usepackage{eccv}

\usepackage{eccvabbrv}
\usepackage{graphicx}
\usepackage{booktabs}
\usepackage{multirow}
\usepackage{tabto}
\usepackage[accsupp]{axessibility} 
\usepackage{hyperref}
\usepackage{orcidlink}
\usepackage{xcolor,colortbl}
\definecolor{boxgrey}{HTML}{F3F3F3}
\newcolumntype{a}{>{\columncolor{boxgrey}}r}
\newcommand{\hlbox}[2]{
  \begin{center}
    \fcolorbox{white}{boxgrey}{
      \parbox{0.9\columnwidth}{\noindent \textbf{#1}. \textit{#2}}
    }
  \end{center}
}

\begin{document}

\title{The Impact of Balancing Real and Synthetic Data on Accuracy and Fairness in Face Recognition} 
\titlerunning{Balancing Real and Synthetic Data in Face Recognition}

\author{Andrea Atzori\orcidlink{0000-0002-6910-206X} \and
Pietro Cosseddu\orcidlink{0009-0006-4998-5164} \and
Gianni Fenu\orcidlink{0000-0003-4668-2476} \and
Mirko Marras\orcidlink{0000-0003-1989-6057}}

\authorrunning{Atzori et al.}

\institute{Department of Mathematics and Computer Science, University of Cagliari, Italy\\
\email{andrea.atzori@ieee.org, p.cosseddu4@studenti.unica.it,\\ fenu@unica.it, mirko.marras@acm.org}
}

\maketitle

\begin{abstract}
  Over the recent years, the advancements in deep face recognition have fueled an increasing demand for large and diverse datasets. Nevertheless, the authentic data acquired to create those datasets is typically sourced from the web, which, in many cases, can lead to significant privacy issues due to the lack of explicit user consent. Furthermore, obtaining a demographically balanced, large dataset is even more difficult because of the natural imbalance in the distribution of images from different demographic groups. In this paper, we investigate the impact of demographically balanced authentic and synthetic data, both individually and in combination, on the accuracy and fairness of face recognition models. Initially, several generative methods were used to balance the demographic representations of the corresponding synthetic datasets. Then a state-of-the-art face encoder was trained and evaluated using (combinations of) synthetic and authentic images. Our findings emphasized two main points: (i) the increased effectiveness of training data generated by diffusion-based models in enhancing accuracy, whether used alone or combined with subsets of authentic data, and (ii) the minimal impact of incorporating balanced data from pre-trained generative methods on fairness (in nearly all tested scenarios using combined datasets, fairness scores remained either unchanged or worsened, even when compared to unbalanced authentic datasets). Source code and data are available at \url{https://cutt.ly/AeQy1K5G} for reproducibility. 
\end{abstract}

\section{Introduction}
\label{sec:intro}

Face Recognition (FR) is one of the most popular biometric tasks. Its applications range from access control to portable devices~\cite{DBLP:journals/cloudcomp/FenuM18,boutros2022pocketnet}. Extremely high levels of accuracy have been achieved thanks to new deep learning architectures~\cite{he2016deep, boutros2023unsupervised}, margin-based losses~\cite{arcface, wang2018cosface, boutros2022elasticface, kim2022adaface} and the availability of large-scale, annotated face datasets~\cite{cao2018vggface2} collected from the Internet. The collection of data from such sources, however, implies that the users involved cannot directly express consent for the use of their data, thereby raising severe ethical concerns. 

The enactment of the General Data Protection Regulation (GDPR)~\cite{GDPR} by the EU in 2018 heightened criticisms regarding privacy issues in this domain. This enactment led to the removal of several databases commonly used in FR~\cite{VGGFace2, umdfaces, msceleb1m} to avert legal complications and cast uncertainty on the future of FR research. The GDPR specifically provides all individuals with the "right to be forgotten" and enforces more rigorous data collection standards. Consequently, there has been a growing focus on synthetic data, which has emerged as a promising substitute for genuine datasets in FR training~\cite{DBLP:journals/ivc/BoutrosSFD23}. This shift has been facilitated by progress in Deep Generative Models (DGMs), which can create synthetic samples by learning the probability distribution of the real ones.

The majority of DGMs are based on Generative Adversarial Networks (GANs) \cite{shoshan2021gan}, Diffusion Models (DMs)~\cite{boutros2023IDiff}~\cite{kim2023dcface}, or, occasionally, hybrid implementations of both~\cite{melzi2023gandiffface}. Presently, FR models using synthetic data typically show a decline in verification accuracy when compared to those trained with authentic data. This performance gap is primarily due to the limited identity discrimination of the training datasets~\cite{Sface} or their low intra-class variance~\cite{boutros2023unsupervised, SynFace}. DMs have gained attention as a plausible alternative to GANs for image synthesis, albeit at the expense of stability and a significant reduction in training performance. Regrettably, several unresolved questions remain regarding the effective combination of authentic and synthetic data to overcome the limitations of both. In a recent study, various combinations of authentic and synthetic data have been used to train FR models and assess the extent to which the use of authentic data can be minimized by introducing synthetic identities, without encountering the aforementioned performance drawbacks~\cite{atzori2024if}. However, the impact of demographically balancing within and among the two sources of data on verification accuracy and fairness has not been considered while training FR models.

This paper aims to investigate the suitability of using combined authentic and synthetic, demographically balanced, training datasets for developing FR models, focusing on both fairness and accuracy. This exploration seeks to determine whether it is possible to simultaneously address performance and fairness concerns while mitigating the privacy-related issues inherent in authentic datasets. By doing so, it may be possible to create accurate and fair FR models with a reduced reliance on authentic data (assuming that synthetic data can be generated without limitation and that a small number of authentic identities can be collected with appropriate user consent). Thus, our contribution is twofold:
\begin{itemize}
    \item We demographically balanced the employed synthetic datasets with respect to the available demographic groups by generating the missing identities using the same methods originally employed, without additional training. The images generated for this study have been made publicly available.    
    \item We investigated whether FR models trained on demographically balanced combinations of authentic and synthetic data could achieve comparable accuracy and fairness to models trained on demographically balanced (and unbalanced) authentic-only data.
\end{itemize}

The rest of the paper is structured as follows. Section \ref{sec:related} discusses recent progress in face recognition methods and synthetic face generation. Section \ref{sec:methodology} then describes the data preparation, model creation and training, and model evaluation adopted in our study. Section \ref{sec:results} examines the differences in verification accuracy and fairness between FR models trained on synthetic and/or authentic data. Finally, Section \ref{sec:conclusions} summarizes our findings and provides directions for future research. Code and data are available at \url{https://cutt.ly/AeQy1K5G}.

\section{Related Work} \label{sec:related}

Our work bridges recent research on fairness in deep face recognition methods and face generation techniques. In this section, we present an overview of both.

\subsubsection{Fairness in Face Recognition.} \label{sec:fairness-ml}

Derived from machine learning literature \cite{DBLP:journals/ipm/BorattoFMM23}, the notions of fairness seek to guarantee fair treatment of individuals across various demographic groups using biometric systems that analyze traits like face, fingerprint, or iris~\cite{TERHORST2020332, 9975333, Kotwal_2024_WACV}. Broadly, demographic fairness is encapsulated by three key concepts: parity, equalized odds, and sufficiency~\cite{article, 10.1145/3457607}. Parity denotes the requirement that the outcome of an FR system should remain unaffected by subject's demographic attributes (such as gender or ethnicity). Equalized odds assert that, regardless of demographic characteristics, the rates of false negatives and false positives should be consistent across demographic groups. Sufficiency implies that the available data must provide sufficient information to ensure accurate and fair results in FR without depending on demographic details. 

Prior work analyzing fairness in face recognition has shown that, on average, women experienced worse performance than men~\cite{biasgenderbio,genderineq,sexist}. 
Further analyses generally attributed this disparity to the fact that female faces were more similar to each other than male faces, as shown in \cite{DBLP:conf/interspeech/MarrasKMF19,genderineq, sexist, DBLP:conf/iir/AtzoriFM23, 10449007}. Notable attention was also paid to factors pertaining to the image (e.g., presence of distortions or noise) or to the face (e.g. presence of make-up or mustache) characteristics~\cite{10007937, ATZORI2022212}. 
For instance, poor performance on dark-skinned or poorly-lit subjects~\cite{10054108} was associated with the fact that the network learns skin-tone-related characteristics already in the top layers. Another demographic dimension whose groups have been shown to be systematically discriminated against is age. Indeed, children's faces were more likely to be badly recognized than those of adults~\cite{ricanek2015review}.
The imbalanced representation of certain groups was also indicated as a possible reason for unfairness~\cite{8638319,9186002}
To counter this, a range of demographically balanced data sets have been created~\cite{inbook,9512390,9025435,demogpairs}. In this study, we analyze the impact of data balancing through the generation of new synthetic identities. Specifically, we are going to analyze how this balancing methodology impacts models trained only on synthetic data and on combined data (authentic and synthetic).

\subsubsection{Synthetic Face Generation.} \label{sec:syntetic-face-gen}
 
Over the last years, several works proposed the use of synthetic data in FR development~\cite{SynFace, boutros2023unsupervised, Digiface1m, Sface, boutros2023IDiff, kim2023dcface, DBLP:journals/ivc/BoutrosSFD23} due to the success of deep generative models in generating high-quality and realistic face images~\cite{GANControl,DiscoFaceGAN,ho2020denoising,improveddiff}. These methods can be categorized as GAN-based~\cite{SynFace,boutros2023unsupervised,Sface,ExFaceGAN}, digital rendering~\cite{Digiface1m}, or diffusion-based~\cite{melzi2023gandiffface,boutros2023IDiff,kim2023dcface}. 

In~\cite{shoshan2021gan}, an architecture based on previous StyleGAN methods~\cite{karras2019style}~\cite{karras2021alias} is presented. Such architecture uses a disentangled latent space to train control encoders that map human-interpretable inputs to suitable latent vectors, thus allowing explicit control of attributes such as pose, age, and expression. By doing so it is then possible to generate new synthetic faces with chosen variations using the controllable attributes. Later, SynFace~\cite{SynFace} proposed to generate synthetic data using an attribute-conditional GAN model, i.e., DiscoFaceGAN~\cite{DiscoFaceGAN}, and perform identity and domain mixup, and SFace~\cite{Sface} analyzed the impact of Style-GAN~\cite{DBLP:conf/cvpr/KarrasLAHLA20} training under class conditional settings and the extent to which transferring knowledge from the pretrained model on authentic data improves the performance of synthetic-based FR. In contrast, ExFaceGAN~\cite{ExFaceGAN} introduced a framework to disentangle identity information within the latent spaces of unconditional GANs,to produce multiple images for any given synthetic identity.

Among methods of digital rendering, DigiFace-1M~\cite{Digiface1m} leveraged facial geometry models, a diverse array of textures, hairstyles, and 3D accessories, along with robust data augmentation techniques during training. However, it comes at a considerable computational cost during the rendering process. DigiFace-1M also proposed combining synthetic and authentic data during FR training to improve the verification accuracy of synthetic-based FR using a small and fixed number of authentic identities.
Recently, IDiff-Face~\cite{boutros2023IDiff} and DCFace~\cite{kim2023dcface} adopted diffusion models to generate synthetic data for FR training, achieving state-of-the-art verification accuracy for synthetic-based FR. Specifically, the former included fuzziness in the identity condition to induce variations in the generated data. Conversely, the latter proposed a two-stage generative framework in which (i) an image of a novel identity using an unconditional diffusion model is generated and an image style from the style bank is selected in order to (ii) be mixed using a dual conditional diffusion model.

Recently, several challenges and competitions have been organized in conjunction with top venues, aiming at promoting privacy-friendly synthetic-based FR development. 
FRCSyn competitions~\cite{Melzi_2024_WACV, deandres2024frcsyn} were organized at WACV and CVPR 2024, aiming to explore the use of synthetic data in FR training and to attract the development of solutions for synthetic-based FR. The challenge considered two main tasks, training FR only with synthetic data and training FR with both synthetic and authentic data. The achieved results of the top-performing solutions from FRCSyn~\cite{Melzi_2024_WACV} competition are further investigated and reported in~\cite{MELZI2024102322}. Also, the SDFR~\cite{shahreza2024sdfr} competition was organized in conjunction with FG 2024, to promote the creation of solutions for synthetic-based FR.

\section{Methodology} \label{sec:methodology}

\begin{figure}[!t]
\centering
\includegraphics[width=.99\textwidth]{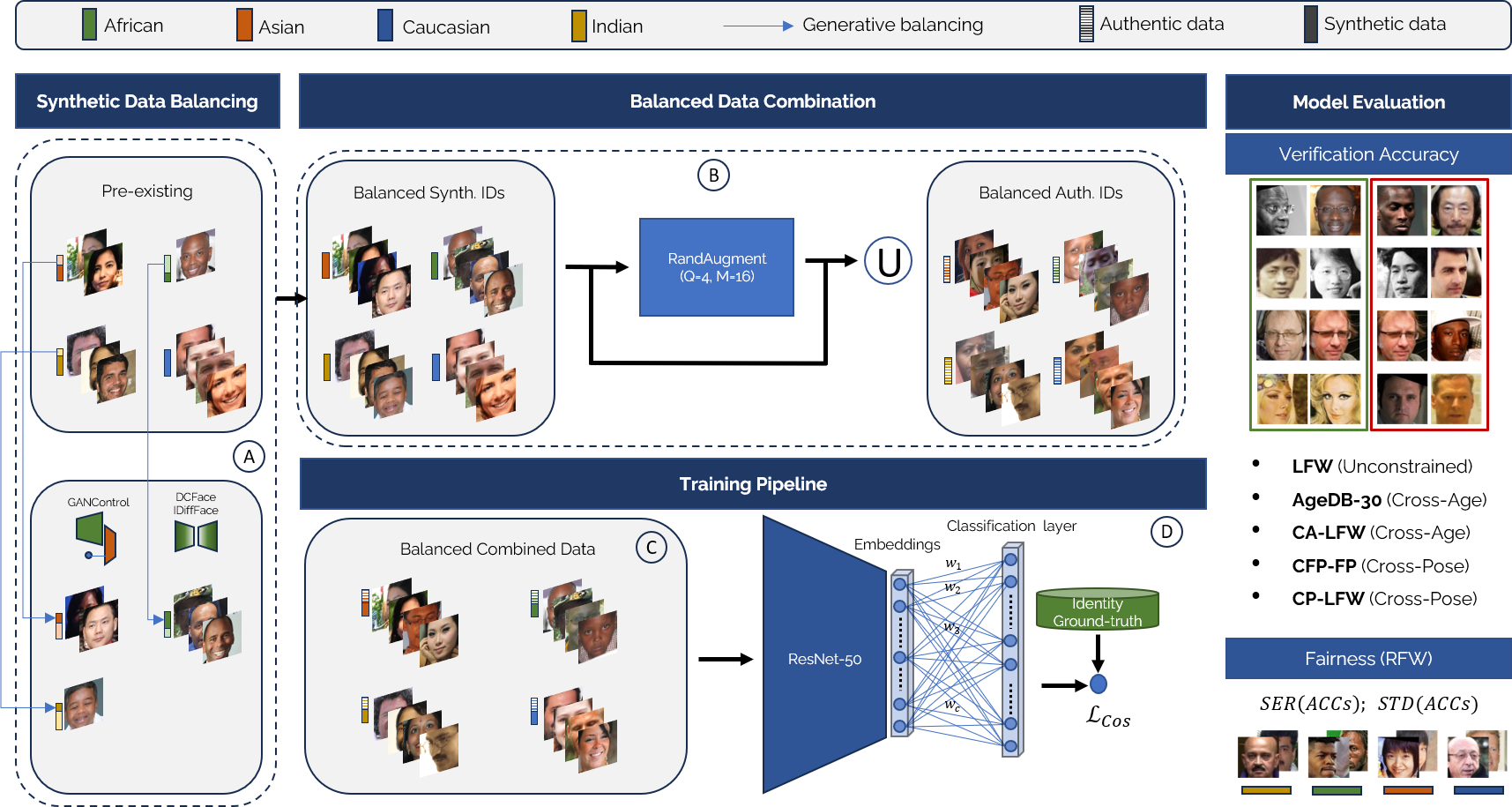}
\caption{In our methodology, firstly synthetic identities are generated to demographically balance the synthetic datasets (\textbf{A}). Subsets of authentic and synthetic data are combined to form the training dataset, with only synthetic data augmented using RandAugment (\textbf{B}). We trained a ResNet50 backbone on the balanced combined data using CosFace loss (\textbf{C, D}). Ultimately, we used different benchmarks to evaluate the FR models' accuracy and fairness, as per our objectives.
}
\label{fig:pipeline}
\end{figure}

This section is dedicated to describing the experimental protocol we followed (Fig.~\ref{fig:pipeline}), including the datasets involved in the experiments, both authentic and synthetic, the training methodologies adopted to combine both types of face data, and the metrics used for model evaluation.

\subsection{Data Preparation} \label{sec:datasets}
For our experiments, we used five different datasets to train the models: two authentic and three synthetic. The datasets were aligned using MTCNN~\cite{zhang2016joint} to extract five facial landmarks, after which all images were resized to 112 $\times$ 112 pixels. Images were normalized to have pixel values between -1 and 1.

\subsubsection{Authentic Datasets.} \label{sec:auth-datasets} 
For authentic face data used to train the FR models, we adopted the well-known BUPT-Balancedface~\cite{wang2020mitigating} and CASIA-WebFace~\cite{yi2014learning} datasets. 
BUPT-Balancedface~\cite{wang2020mitigating} consists of 1.3M images from 28K identities and is annotated with both ethnicity and identity labels. Its ethnicity annotations include four demographic groups: African, Asian, Caucasian, and Indian, with 7K identities and approximately 300K images each.
Conversely, CASIA-WebFace~\cite{yi2014learning} consists of 0.5M images of 10K identities. It is worth noting that this dataset was included in the experiments as a reference, despite not being demographically balanced. In~\cite{kim2023dcface}, it is reported a demographic distribution of 63.4\% Caucasian, 14.4\% Asian, 7.4\% African, 7.2\% Indian, and 7.4\% Others.

\subsubsection{Synthetic Datasets.} \label{sec:syn-datasets}
The synthetic datasets were generated using three methods: one GAN-based and two diffusion-based. These datasets are derived from ExFaceGAN~\cite{ExFaceGAN, GANControl}, DCFace~\cite{kim2023dcface}, and IDiff-Face Uniform (25\% CPD)~\cite{boutros2023IDiff}. Each dataset contains 0.5M images from 10K identities, with 50 images per identity.

The first synthetic dataset was generated via the pretrained GAN-Control~\cite{GANControl} generator, which was trained on the FFHQ dataset~\cite{karras2019style} and improved with an identity disentanglement approach~\cite{ExFaceGAN}.
The second synthetic dataset was generated via DCFace~\cite{kim2023dcface}, which is based on a two-stage diffusion model. In the first stage, a high-quality face image of a novel identity is generated using unconditional diffusion models~\cite{ho2020denoising} trained on FFHQ~\cite{karras2019style}, with the image style randomly selected from a style bank. In the second stage, the generated images and styles from the first stage are combined using a dual conditional diffusion model~\cite{ho2020denoising} trained on CASIA-WebFace~\cite{yi2014learning} to produce an image with a specific identity and style.
Finally, the third synthetic dataset was generated via IDiff-Face~\cite{boutros2023IDiff}, a novel approach based on conditional latent diffusion models for synthetic identity generation with realistic identity variations for FR training. IDiff-Face is trained in the latent space of a pretrained autoencoder~\cite{rombach2022high} and conditioned on identity contexts (i.e., feature representations extracted using a pretrained FR model, namely ElasticFace~\cite{boutros2022elasticface}).

\begin{table}[!t]
\centering
\resizebox{1.\linewidth}{!}{%
\begin{tabular}{l|r|r|r|r}
\hline \textbf{Generative Method} \tabto{1cm} & \textbf{\# Caucasian IDs} & \textbf{\# Indian IDs} & \textbf{\# Asian IDs} & \textbf{\# African IDs} \\
\hline ExFaceGAN~\cite{ExFaceGAN} & 6,218 & 1,973 & 1,668 & 141 \\ 
\hline DCFace~\cite{kim2023dcface} & 8,290 & 887 & 571 & 252 \\ 
\hline IDiff-Face~\cite{boutros2023IDiff} & 7,464 & 1,090 & 915 & 580 \\ 
\hline
\end{tabular}}
\caption{Demographic representation within the synthetic datasets used in our study.}
\label{table:datasets-ethnicity}
\end{table}

\subsubsection{Data Sampling and Balancing.} \label{syn-balancing}
The authentic dataset employed in the majority of our experiments, BUPT-Balancedface, was already demographically balanced, containing an equal number of identities across the four demographic groups. For our experiments, we required 5K unique, demographically balanced identities, aiming for a total of 1,250 identities per demographic group. To achieve this and reduce the randomness in our experiments, we randomly sampled identities ten times, with each iteration including 5K demographically balanced identities from BUPT-Balancedface. We denote the best-performing iteration as BUPT$_{sub}$, which was used in subsequent experiments. The average results across all iterations are referred to as BUPT$_{avg}$. Similarly, the average verification accuracy across the ten iterations from CASIA-WebFace is denoted as WF$_{avg}$, while the best-performing iteration is referred to as WF$_{sub}$.

The synthetic datasets were unbalanced towards the Caucasian group, as determined by labeling all the data using a ResNet18~\cite{he2016deep} backbone trained on BUPT-BalancedFace~\cite{wang2020mitigating} to predict the ethnicity label of each identity. The inferred ethnicity pseudo-labels are reported in Table~\ref{table:datasets-ethnicity}. For our experiments, we required 5K unique, demographically balanced identities, aiming for a total of 1,250 identities per demographic group in each synthetic dataset.
To achieve this, we (i) randomly sampled 1,250 identities (or the available number, if fewer) from the synthetic datasets and (ii) generated new identities for each demographic group until reaching our targets by guiding the generation process with the above-mentioned ResNet18~\cite{he2016deep} backbone. For each synthetic dataset, the additional identities were generated using the pre-trained models made publicly available by the original authors without further training. 
We denote the synthetic subsets sampled in the first step as GC$_{sub}$, DC$_{sub}$, and IDF$_{sub}$, and the ones generated in the second step as GC$_{gen}$, DC$_{gen}$, and IDF$_{gen}$, using GANControl, DCFace, and IDiff-Face, respectively. Finally, the synthetic, demographically balanced datasets, each comprising 5K identities and derived from the union of the two respective datasets for each method, are referred to as GC$_{bal}$, DC$_{bal}$, and IDF$_{bal}$ for the sake of clarity. 

\subsubsection{Training Data Combination.} \label{sec:combinations}
We trained FR models using combinations of authentic and synthetic data. The authentic subset involved in each combination was always BUPT$_{sub}$, which consists of 5K identities and is balanced across demographic groups. This subset was then combined with each of the three synthetic, demographically balanced subsets (GC$_{bal}$, DC$_{bal}$, IDF$_{bal}$), all of which have the same demographic distribution and the same number of identities (5K).

\subsection{Model Creation and Training} \label{sec:training}
To train all the FR models we relied on the widely used ResNet50~\cite{he2016deep} as the backbone and CosFace~\cite{wang2018cosface} as the loss function. The latter is defined as:
\begin{equation}
    L_{\text{CosFace}} = \frac{1}{N} \sum_{i \in N} -\log \frac{e^{s (\cos(\theta_{y_i}) - m)}}{e^{s (\cos(\theta_{y_i}) - m)} + \sum_{j=1, j \neq y_i}^{c} e^{s \cos(\theta_j)}}
\end{equation}
where \( c \) is the number of classes (identities), \( N \) is the batch size, \( m \) is the margin penalty applied on the cosine angle $cos(\theta_{y_i})$ between the feature representation \( x_i \) of the sample \( i \) and its class center \( y_i \), \( s \) is the scale parameter. In all the conducted experiments, the margin \( m \) is set to 0.35 and the scale parameter \( s \) to 64, following~\cite{wang2018cosface}. During the training, we employed Stochastic Gradient Descend (SGD) as an optimizer with an initial learning rate of 0.1. The learning rate is divided by 10 at epochs 22, 30 and 40. In total, the models are trained for 40 epochs using 256 as batch size.
During the training, we also employed data augmentation techniques, following RandAugment~\cite{cubuk2020randaugment}. Its augmentation space includes color and geometric transformations such as horizontal flipping, sharpness adjusting, and translation of the \(x \) and \(y\) axes. RandAugment includes two hyper-parameters, \(Q\) and \(M\), to select the number of operations \(Q\) and the magnitude \(M\) of each transformation. In our experiments, \(M\) and \(Q\) were set to 16 and 4, as in~\cite{atzori2024if} and~\cite{boutros2023unsupervised}. Further details are provided in the code repository.

\subsection{Model Evaluation} \label{sec:metrics}
We evaluated the trained FR models in terms of verification accuracy on several well-known benchmarks, accompanying the following datasets: LFW~\cite{huang2008labeled}, CFP-FP~\cite{sengupta2016frontal}, CFP-FF~\cite{sengupta2016frontal}, AgeDB-30~\cite{moschoglou2017agedb}, CA-LFW~\cite{zheng2017cross}, CP-LFW~\cite{sengupta2016frontal} and RFW~\cite{wang2019racial}. The latter has also been used to assess the fairness of the trained FR models. Results for all benchmarks are reported as verification accuracy in percentage, thus adhering to their official, original evaluation protocol.

In order to assess the fairness of the models, we computed the standard deviation (STD) and the Skewed Error Ratio (SER) on the verification accuracy of the four sub-groups composing the RFW benchmark, with each sub-group composed of 6K mated and 6K non-mated verification pairs. Specifically, error skewness is computed as the ratio of the highest error rate to the lowest error rate among different demographic groups. Formally:

\begin{equation}\label{eq:ser}
    SER = \frac{\max_a \text{Err}(a)}{\min_b \text{Err}(b)}
\end{equation}

where \(a\) and \(b\) are different demographic groups. In this context, a higher error skewness indicates that the model has a substantial discrepancy in accuracy between the best and worst performing demographic groups, and is thus less fair. On the other hand, the metric based on the standard deviation is defined as:

\begin{equation}\label{eq:standard-deviation}
    STD = \sqrt{\frac{1}{N} \sum_{i=1}^{N} (E_i - \bar{E})^2}
\end{equation}

where \(E_i\) is the error rate for demographic group \(i\), \(N\) is the total number of demographic groups, and \(\bar{E}\) is the mean error rate across all groups. A higher standard deviation indicates that the model has substantially different verification accuracies across demographic groups and is therefore less fair.

\section{Experimental Results}  \label{sec:results}

Our experiments initially aimed to assess whether an FR model trained on a demographically balanced synthetic dataset could achieve competitive accuracy compared to an FR model trained on an authentic dataset with the same number of identities and demographic representation (Section~\ref{RQ1}). Subsequently, we explored the impact on verification accuracy by training FR models on combined synthetic and authentic data (Section~\ref{RQ2}) and investigated the impact on the fairness of each setting involved in our study (Section~\ref{RQ3}).

\subsection{RQ1: Accuracy with Separate Synthetic and Real Data Training} \label{RQ1}
In a first analysis, we assessed whether an FR model trained on a demographically balanced synthetic dataset can achieve competitive accuracy compared to an FR model trained on an authentic dataset with the same number of identities and demographic representation. To this end, Tab. \ref{tab:results-acc-NOAUG-5k} (without data augmentation) and \ref{tab:results-acc-AUG-5k} (with data augmentation) present the accuracy of the FR models trained on authentic and synthetic datasets, separately, with 5K identities.

\begin{table}[!b]
    \centering
    \resizebox{\textwidth}{!}{%
    \begin{tabular}{l|r|r|r|r|r|r|r|r}
        \hline
        \textbf{Train Data} & \textbf{Id/Img.} & \textbf{LFW} & \textbf{CFP-FP} & \textbf{CFP-FF} & \textbf{AgeDB-30} & \textbf{CA-LFW} & \textbf{CP-LFW} & \textbf{Avg.} \\
        \hline
        BUPT$_{avg}$ & 5K/42 & 92.98 & 72.18 & 92.38 & 75.78 & 80.78 & 69.45 & 80.59 \\ 
        \hline
        WF$_{avg}$ & 5K/46 & \textbf{98.91} & \textbf{92.15} & \textbf{98.92} & \textbf{91.46} & \textbf{91.76} & \textbf{85.91} & \textbf{93.19} \\ 
        \hline
        \hline
        BUPT$_{sub}$  & 5K/42 & 92.98 & 71.18 & 92.38 & 75.78 & 80.78 & 69.45 & 80.59 \\ 
        \hline
        WF$_{sub}$ & 5K/46 & \textbf{98.95} & \textbf{92.27} & \textbf{98.94}  & \textbf{91.40}  & \textbf{91.93} & \textbf{86.22} & \textbf{93.28} \\ 
        \hline
        \hline
        GC$_{bal}$  & 5K/47 & 86.96 & 71.67 & 85.52 & 59.56 & 71.83 & 65.56 & 73.65 \\
        \hline
        DC$_{bal}$  & 5K/48 & \textbf{97.23} & \textbf{83.45} & \textbf{97.58} & \textbf{85.95} & \textbf{89.11} & \textbf{78.06} & \textbf{88.50} \\
        \hline
        IDF$_{bal}$ & 5K/47 & 96.50 & 77.44 & 95.15 & 80.10 & 88.05 & 76.55 & 85.63 \\
        \hline
    \end{tabular}%
    }
    \vspace{1mm}\caption{Verification accuracy of FR models trained on 5K identities \emph{without data augmentation}. The results are reported for models trained: (i) only on authentic data, averaged across 10 iterations, (ii) only on authentic data, for the best performing iteration, and (iii) only on synthetic, demographically balanced data. The best results for each group are highlighted in bold.}
    \label{tab:results-acc-NOAUG-5k}
\end{table}
\begin{table}[!b]
    \centering
    \resizebox{\textwidth}{!}{%
    \begin{tabular}{l|r|r|r|r|r|r|r|r}
        \hline
        \textbf{Train Data} & \textbf{Id/Img.} & \textbf{LFW} & \textbf{CFP-FP} & \textbf{CFP-FF} & \textbf{AgeDB-30} & \textbf{CA-LFW} & \textbf{CP-LFW} & \textbf{Avg.} \\
        \hline
        BUPT$_{sub}$ & 5K/42 & 92.90 & 75.65 & 93.22 & 76.78 & 81.28 & 70.88 & 81.72 \\ 
        \hline
        WF$_{sub}$  & 5K/46 & \textbf{98.91} & \textbf{92.17} & \textbf{99.02} & \textbf{91.30} & \textbf{92.21} & \textbf{86.03} & \textbf{93.27} \\ 
        \hline
        \hline
        GC$_{bal}$  & 5K/47 & 93.68 & 75.38 & 91.64 & 79.03 & 82.31 & 72.25 & 82.31 \\
        \hline
        DC$_{bal}$   & 5K/48 & \textbf{97.45} & \textbf{86.42} & \textbf{97.32} & \textbf{87.01} & \textbf{89.33} & \textbf{80.00} & \textbf{89.52} \\
        \hline
        IDF$_{bal}$  & 5K/47 & 96.91 & 80.82 & 95.24 & 82.56 & 88.00 & 77.53 & 86.88 \\
        \hline
    \end{tabular}%
    }
    \vspace{1mm}
    \vspace{1mm}\caption{Verification accuracy of FR models trained on 5K identities \emph{with data augmentation}. The results are reported for models trained: (i) only on authentic data, averaged across 10 iterations, (ii) only on authentic data, for the best performing iteration, and (iii) only on synthetic, demographically balanced data. The best results for each group are highlighted in bold.}
    \label{tab:results-acc-AUG-5k}
\end{table}

In our investigation, models trained exclusively on authentic data without the application of data augmentation (Tab. \ref{tab:results-acc-NOAUG-5k}, first two groups) consistently exhibited superior verification accuracy when trained on subsets of the CASIA-WebFace dataset. This trend was observed both when considering average performance across iterations (WF$_{avg}$) and the best iteration outcomes (WF$_{sub}$), with these models showing an approximately 15\% improvement in verification accuracy w.r.t. the respective one trained on demographically balanced subsets of BUPT (BUPT$_{avg}$ and BUPT$_{sub}$). In contrast, among the models trained solely on synthetic images (Tab. \ref{tab:results-acc-NOAUG-5k}, third group), the model trained on the DC$_{bal}$ subset achieved the highest verification accuracy across all evaluation benchmarks. Specifically, the latter model outperformed the one trained on the IDF$_{bal}$ subset by an average of 3.35\% and the one trained on the GC$_{bal}$ subset by a substantial 20.16\%. Interestingly, we observed a pronounced accuracy degradation of the FR model trained on the GC$_{bal}$ subset, when evaluated on cross-age benchmarks (AgeDB-30 and CA-LFW columns). For instance, compared to the models trained on DC$_{bal}$, GC$_{bal}$-trained models exhibited a 30.66\% reduction on AgeDB-30 and a 19.39\% decrease on CA-LFW.
Comparing between models trained with the two different types of sources separately (authentic and synthetic), models trained exclusively on synthetic data from DC$_{bal}$ and IDF$_{bal}$ generally achieved better verification accuracy compared to models trained on the authentic, demographically-balanced BUPT$_{sub}$ subset. Specifically, the model trained on the DC$_{bal}$ subset obtained 9.82\% higher average verification accuracy, while training on the IDF$_{bal}$ subset led to a 6.25\% gain, on average. Despite the promising results achieved by training an FR model on the best-performing synthetic dataset (DC$_{bal}$), a substantial gap of 5.40\% in average verification accuracy remains when compared to the best-performing authentic dataset (CASIA$_{sub}$).

The impact of data augmentation on models trained solely on synthetic data (Tab. \ref{tab:results-acc-AUG-5k}, second group) was notably pronounced, especially for GC$_{bal}$. The model trained on the latter, augmented subset, led to an average accuracy improvement of 11.75\% compared to the corresponding model trained without augmentation. This improvement was particularly pronounced on cross-age benchmarks, with a remarkable 34.42\% increase in verification accuracy on AgeDB-30 and a 15.59\% increase on CA-LFW. Furthermore, all the models trained on synthetic datasets still reported higher verification accuracy compared to those trained on the balanced, augmented authentic data (BUPT$_{sub}$), with the smallest improvement observed while training on GC$_{bal}$ (0.72\%) and the highest improvement measured while training on DC$_{bal}$ (9.54\%). On the other hand, adding data augmentation to the training pipeline of models trained exclusively on authentic data (Tab. \ref{tab:results-acc-AUG-5k}, first group) resulted in only marginal improvements, where the maximum increase in accuracy was limited to 1.40\% (BUPT$_{sub}$). Comparing results obtained by training an FR model on DC$_{bal}$ and CASIA$_{sub}$ while applying data augmentation, it can be noted that the accuracy gap between training on authentic and synthetic data is reduced (4.19\%) with respect to the gap obtained by training on the same datasets without data augmentation.

\hlbox{RQ1}{Models trained on synthetic data, especially when supplemented with data augmentation, tend to get closer (CASIA-WebFace) or even outperform (BUPT) those trained on authentic (balanced) data, with the highest gains observed in cross-age tasks. The integration of data augmentation substantially mitigated performance degradation in models trained on the GC$_{bal}$ subset, especially concerning cross-age benchmarks.}

\subsection{RQ2: Accuracy with Combined, Balanced Training Data} \label{RQ2}
In a second analysis, we explored the impact on verification accuracy by training FR models using a combination of synthetic and authentic data. To this end, Tab. \ref{tab:results-acc-NOAUG-10k} (without data augmentation) and \ref{tab:results-acc-AUG-10k} (with data augmentation) report the verification accuracy of FR models trained on datasets (either entirely authentic or combined), each composed of 10K identities.

Models trained exclusively on authentic data without data augmentation (Tab. \ref{tab:results-acc-NOAUG-10k}, first group) highlighted (again) a substantial gap in verification accuracy between the model trained on CASIA-WebFace and the one trained on BUPT$_{10K}$, with a 14.32\% difference. FR models trained on a demographically balanced combination of synthetic and authentic data without data augmentation (Tab. \ref{tab:results-acc-NOAUG-10k}, second group) consistently outperformed the baseline model trained solely on BUPT$_{10K}$. Specifically, these models obtained 4.04\% (BUPT$_{sub}$ $\cup$ GC$_{bal}$), 7.36\% (BUPT$_{sub}$ $\cup$ IDF$_{bal}$), and 9.71\% (BUPT$_{sub}$ $\cup$ DC$_{bal}$) higher verification accuracy. Notably, when training an FR model on the combined BUPT$_{sub}$ $\cup$ GC$_{bal}$ dataset without data augmentation, the accuracy degradation identified on cross-age benchmarks in the previous subsection was not observed, suggesting that the inclusion of a balanced authentic data subset (BUPT$_{sub}$) effectively mitigates these issues. The best verification accuracy across all benchmarks was achieved by models trained on the combined dataset including DC$_{bal}$ as the synthetic component (BUPT$_{sub}$ $\cup$ DC$_{bal}$). This model showed an average accuracy increase of 1.18\% over the one trained on BUPT$_{sub}$ $\cup$ IDF$_{bal}$ and 5.44\% over the one trained on BUPT$_{sub}$ $\cup$ GC$_{bal}$. Comparing results obtained by training an FR model on BUPT$_{sub}$ $\cup$ DC$_{bal}$ and CASIA-WebFace, it can be noted that while the accuracy gap between training on authentic and combined (authentic and synthetic) data is reduced, it remains remarkable, with a 4.20\% difference.

\begin{table}[!t]
    \centering
    \resizebox{\textwidth}{!}{%
    \begin{tabular}{l|r|r|r|r|r|r|r|r}
        \hline
        \textbf{Train Data} & \textbf{Id/Img.} & \textbf{LFW} & \textbf{CFP-FP} & \textbf{CFP-FF} & \textbf{AgeDB-30} & \textbf{CA-LFW} & \textbf{CP-LFW} & \textbf{Avg.} \\
        \hline
        BUPT$_{10K}$  & 10K/42 & 95.55 & 74.48 & 95.88 & 79.95 & 85.28 & 69.61 & 83.42 \\ 
        \hline
        CASIA-WebFace  & 10K/46 &  \textbf{99.46} & \textbf{95.12} & \textbf{99.51} & \textbf{94.61} & \textbf{93.90} & \textbf{83.63} & \textbf{95.37} \\ 
        \hline
        \hline
        BUPT$_{sub}$ $\cup$ GC$_{bal}$ & 10K/45 &  97.25 & 79.94 & 95.57 & 82.93 & 86.30 & 78.40 & 86.79 \\
        \hline
        BUPT$_{sub}$ $\cup$ DC$_{bal}$  & 10K/45 &  \textbf{98.55} & \textbf{87.72} & \textbf{98.64} & \textbf{90.33} & \textbf{91.86} & \textbf{82.20} & \textbf{91.52} \\
        \hline
        BUPT$_{sub}$ $\cup$ IDF$_{bal}$ & 10K/45 & 98.18 & 83.82 & 97.62 & 86.83 & 91.26 & 81.06 & 89.86 \\
        \hline
    \end{tabular}%
    }
    \vspace{1mm}
    \caption{Verification accuracy of FR models trained on 10K identities \emph{without data augmentation}. Results are reported for models (i) trained only on authentic data and (ii) trained on demographically balanced, combined data. BUPT$_{10K}$ denotes a demographically balanced subset of 10K identities (2.5K per group) from BUPT-Balancedface, whereas CASIA-WebFace was not demographically balanced. Best results for each group are highlighted in bold.}
    \label{tab:results-acc-NOAUG-10k}
\end{table}
\begin{table}[!t]
    \centering
    \resizebox{\textwidth}{!}{%
    \begin{tabular}{l|r|r|r|r|r|r|r|r}
        \hline
        \textbf{Train Data} & \textbf{Id/Img.} & \textbf{LFW} & \textbf{CFP-FP} & \textbf{CFP-FF} & \textbf{AgeDB-30} & \textbf{CA-LFW} & \textbf{CP-LFW} & \textbf{Avg.} \\
        \hline
        BUPT$_{10K}$  & 10K/42 &  93.38  & 73.55 & 94.45 & 76.86 & 82.70 & 70.65 & 81.19 \\
        \hline
        CASIA-WebFace  & 10K/46 &  \textbf{99.50}  & \textbf{95.45} & \textbf{99.40} & \textbf{94.15}  & \textbf{93.15}  & \textbf{89.95} & \textbf{95.26}\\ 
        \hline
        \hline
        BUPT$_{sub}$ $\cup$ GC$_{bal}$  & 10K/45 &  97.36 & 81.24 & 95.70 & 84.28 & 88.25 & 78.36 & 87.56 \\
        \hline
        BUPT$_{sub}$ $\cup$ DC$_{bal}$  & 10K/45 &  \textbf{98.45} & \textbf{89.62} & \textbf{98.55} & \textbf{90.20} & \textbf{91.55} & \textbf{83.85} & \textbf{92.00}\\
        \hline
        BUPT$_{sub}$ $\cup$ IDF$_{bal}$  & 10K/45 &  98.13 & 84.91 & 97.31 & 87.23 & 90.66 & 81.50 & 89.97 \\
        \hline
    \end{tabular}%
    }
    \vspace{1mm}
    \caption{Verification accuracy of FR models trained on 10K identities \emph{with data augmentation}. Results are reported for models (i) trained only on authentic data and (ii) trained on demographically balanced, combined data. BUPT$_{10K}$ denotes a demographically balanced subset of 10K identities (2.5K per group) from BUPT-Balancedface, whereas CASIA-WebFace was not demographically balanced. Best results for each group are highlighted in bold.}
    \label{tab:results-acc-AUG-10k}
\end{table}

FR models trained with data augmentation only on authentic data (Tab. \ref{tab:results-acc-AUG-10k}, first group) showed slight decreases in verification accuracy w.r.t. the non-augmented counterpart, with degradations of 2.67\% (BUPT$_{10K}$) and 0.11\% (CASIA-WebFace). Conversely, while the impact of data augmentation on models trained on combined synthetic and authentic data (Tab. \ref{tab:results-acc-AUG-10k}, second group) was generally positive, the improvement was minimal. The models reported an increase in average verification accuracy of 0.88\% when trained on BUPT$_{sub}$ $\cup$ GC$_{bal}$, 0.45\% on BUPT$_{sub}$ $\cup$ IDF$_{bal}$, and 0.52\% on BUPT$_{sub}$ $\cup$ DC$_{bal}$. As previously observed, including data augmentation in the training pipeline positively affects the verification accuracy gap observed when comparing the results of the FR model trained on the best-performing authentic (CASIA-WebFace) and combined (BUPT$_{sub}$ \(\cup\) DC$_{bal}$) datasets, leading to a reduced 3.54\% difference.

\hlbox{RQ2}{Combining demographically balanced synthetic and authentic data can improve verification accuracy compared to training exclusively on authentic data, particularly in the absence of data augmentation. The inclusion of balanced authentic data effectively mitigates potential cross-age accuracy degradation. Data augmentation provides modest changes.}

\subsection{RQ3: Fairness with Combined, Balanced Training Data} \label{RQ3}
In the third and final analysis, we investigated the impact on fairness of each setting involved in our study. To this end, Tab. \ref{tab:results-fairness-NOAUG} (without data augmentation) and \ref{tab:results-fairness-AUG} (with data augmentation) present the verification accuracy for each demographic group, as well as the standard deviation (STD) and the skewed error ratio (SER) on the RFW dataset's benchmark used to evaluate the fairness of FR models. Higher values of STD and SER indicate a higher level of unfairness.

\begin{table}[!b]
    \centering
    \resizebox{\textwidth}{!}{%
    \begin{tabular}{l|r|r|r|r|r|r|r|r}
        \hline
        \textbf{Train Data}  & \textbf{Id/Img.} & \textbf{African($\uparrow$)}  & \textbf{Asian($\uparrow$)} & \textbf{Caucasian($\uparrow$)} & \textbf{Indian($\uparrow$)}  & \textbf{Avg.($\uparrow$)} & \textbf{STD($\downarrow$)} & \textbf{SER($\downarrow$)}\\
        \hline
        BUPT$_{10K}$  & 10K/42 & 72.68 & 75.93 & 79.35 & 79.15 & 76.77 & \textbf{2.72} & 1.09\\ 
        \hline
        CASIA-WebFace  & 10K/46 & \textbf{87.45} & \textbf{86.31} & \textbf{93.95} & \textbf{89.45}  & \textbf{89.29} & 2.91 & \textbf{1.08}\\ 
        \hline
        \hline
        BUPT$_{sub}$ $\cup$ GC$_{bal}$  & 10K/45 & 73.08 & 76.10 & 79.73 & 77.50 & 76.60 & 2.41 & 1.09 \\
        \hline
        BUPT$_{sub}$ $\cup$ DC$_{bal}$  & 10K/45 & 78.58 & 79.96 & \textbf{86.38} & 83.61 & 82.13 & 3.06 & 1.09\\
        \hline
        BUPT$_{sub}$ $\cup$ IDF$_{bal}$  & 10K/45 & \textbf{79.48} & \textbf{82.00} & 85.83 & \textbf{83.81} & \textbf{82.78} & \textbf{2.33} & \textbf{1.07}\\
        \hline
        \hline
        BUPT$_{avg}$ & 5K/42 & 68.45 & 72.37 & 75.45 & 74.38 & 72.66 & \textbf{2.67} & \textbf{1.10}\\ 
        \hline
        WF$_{avg}$ & 5K/42 &  \textbf{80.76} & \textbf{80.83}  & \textbf{89.71} & \textbf{84.63} & \textbf{83.98} & 3.65 & 1.11\\ 
        \hline
        \hline
        BUPT$_{sub}$  & 5K/42 & 68.06 & 71.66 & 75.38 & 74.36 & 72.37 & \textbf{2.83} & \textbf{1.10}\\ 
        \hline
        WF$_{sub}$  & 5K/46 & \textbf{81.31} & \textbf{80.61} & \textbf{89.78} & \textbf{84.60} & \textbf{84.07} & 3.62 & 1.11\\ 
        \hline
        \hline
        GC$_{bal}$  & 5K/47 & 57.95 & 64.41 & 66.01 & 63.65 & 63.00 & \textbf{3.04} & 1.13\\
        \hline
        DC$_{bal}$   & 5K/47 & 69.98 & 74.80 & \textbf{82.21} & 77.81 & 76.20 & 4.45 & 1.17\\
        \hline
        IDF$_{bal}$  & 5K/47 & \textbf{71.96} & \textbf{76.90} & 81.23 & \textbf{77.95} & \textbf{77.01} & 3.32 & \textbf{1.12}\\
        \hline
    \end{tabular}%
    }
    \vspace{1mm}
    \caption{Fairness on RFW demographic groups \emph{without data augmentation}. We report the verification accuracy, standard deviation, and skewed error for FR models trained: (i) only on authentic data (10K), (ii) on demographically balanced, combined data (10K), (iii) only on authentic data, averaged across ten iterations (5K), (iv) only on authentic data, on the best iteration (5K), and (v) only on synthetic, demographically balanced data. BUPT$_{10K}$ denotes a demographically balanced subset of 10K identities (2.5K per group) from BUPT-Balancedface, whereas CASIA-WebFace was not demographically balanced. Best results for each group are highlighted in bold.}
    \label{tab:results-fairness-NOAUG}
\end{table}

\begin{table}[!t]
    \centering
    \resizebox{\textwidth}{!}{%
    \begin{tabular}{l|r|r|r|r|r|r|r|r}
        \hline
        \textbf{Train Data}  & \textbf{Id/Img.} & \textbf{African($\uparrow$)}  & \textbf{Asian($\uparrow$)} & \textbf{Caucasian($\uparrow$)} & \textbf{Indian($\uparrow$)}  & \textbf{Avg.($\uparrow$)} & \textbf{STD($\downarrow$)} & \textbf{SER($\downarrow$)}\\ 
        \hline
        BUPT$_{10K}$  & 10K/42 & 67.76  & 73.11 & 76.80 & 72.20 & 73.47 & 3.21 & 1.13\\
        \hline
        CASIA-WebFace  &  10K/46 & \textbf{87.00} & \textbf{85.53} & \textbf{93.51}  & \textbf{88.90}  & \textbf{88.73} & \textbf{3.00} &  \textbf{1.09}\\ 
        \hline
        \hline
        BUPT$_{sub}$ $\cup$ GC$_{bal}$  &  10K/45 & 71.35 & 77.18 & 80.53 & 77.81 & 76.72 & 3.34 & 1.12\\
        \hline
        BUPT$_{sub}$ $\cup$ DC$_{bal}$  &  10K/45 & \textbf{79.68} & 81.26 & \textbf{87.58} & \textbf{84.56} & \textbf{83.27} & \textbf{3.04} & \textbf{1.09}\\
        \hline
        BUPT$_{sub}$ $\cup$ IDF$_{bal}$  &  10K/45 & 76.65 & \textbf{82.16} & 85.75 & 83.41 & 82.49 & 3.34 & 1.11\\
        \hline
        \hline
        BUPT$_{sub}$  &  5K/42 & 64.86 & 70.91 & 74.60 & 73.35 & 70.93 & \textbf{3.74} & 1.14\\ 
        \hline
        WF$_{sub}$  &  5K/46 & \textbf{80.76} & \textbf{80.08} & \textbf{89.50} & \textbf{84.81} & \textbf{83.79} & 3.76 & \textbf{1.11}\\ 
        \hline
        \hline
        GC$_{bal}$  &  5K/47 & 63.21 & 71.46 & 73.91 & 71.68 & 70.00 & 4.07 & 1.16\\
        \hline
        DC$_{bal}$   &  5K/48 & 71.78 & 75.95 & \textbf{83.55} & \textbf{79.10} & 77.59 & 4.30 & 1.16\\
        \hline
        IDF$_{bal}$  &  5K/47 & \textbf{73.36} & \textbf{77.25} & 81.88 & 78.80 & \textbf{77.82} & \textbf{3.06} & \textbf{1.11}\\
        \hline
    \end{tabular}%
    }
    \vspace{1mm}
    \caption{Fairness on RFW demographic groups \emph{with data augmentation}. We report the verification accuracy, standard deviation, and skewed error for FR models trained: (i) only on authentic data (10K), (ii) on demographically balanced, combined data (10K), (iii) only on authentic data, averaged across ten iterations (5K), (iv) only on authentic data, on the best iteration (5K), and (v) only on synthetic, demographically balanced data. BUPT$_{10K}$ denotes a demographically balanced subset of 10K identities (2.5K per group) from BUPT-Balancedface, whereas CASIA-WebFace was not demographically balanced. Best results for each group are highlighted in bold.}
    \label{tab:results-fairness-AUG}
\end{table}

On the RFW benchmark, models trained exclusively on authentic data without data augmentation (Tab. \ref{tab:results-fairness-NOAUG}, first group) revealed that training on the balanced dataset (BUPT$_{10K}$) led to lower verification accuracy compared to CASIA-WebFace, with a notable gap of 16.19\%. Although training on BUPT$_{10K}$ led to a slight improvement in terms of fairness, as indicated by a 6.52\% reduction in STD, it also showed a slight negative impact on SER. A similar trend was observed when training FR models on smaller subsets with 5K identities, BUPT$_{sub}$ and WF$_{sub}$ (Tab. \ref{tab:results-fairness-NOAUG}, third and fourth groups), where the balanced subset showed marginally better fairness but still under-performed in verification accuracy.

The results achieved by training FR models on synthetic balanced subsets (Tab. \ref{tab:results-fairness-NOAUG}, second and fifth groups), either alone or in combination with BUPT$_{sub}$, slightly diverged from previous observations. Among the models trained solely on synthetic data (Tab. \ref{tab:results-fairness-NOAUG}, second group), the model trained on IDF$_{bal}$ achieved the highest average verification accuracy, outperforming those trained on DC$_{bal}$ by 1.06\% and on GC$_{bal}$ by 22.23\%. Additionally, the model trained on IDF$_{bal}$ reported the best SER (1.02), while the model trained on GC$_{bal}$ achieved the lowest STD. Training on combined balanced datasets (Tab. \ref{tab:results-fairness-NOAUG}, fifth group) led to similar patterns. The model trained on BUPT$_{sub}$ $\cup$ IDF$_{bal}$ exhibited the best average accuracy across demographic groups (82.78\%) and the lowest SER and STD (1.07 and 2.33, respectively). Models trained on the other combined datasets (BUPT$_{sub}$ $\cup$ GC$_{bal}$ and BUPT$_{sub}$ $\cup$ DC$_{bal}$) reported a SER of 1.09, but differences were noted in average STD and verification accuracy. Specifically, the model trained on BUPT$_{sub}$ $\cup$ DC$_{bal}$ achieved 8.06\% higher accuracy but a worse STD (-26.97\%) compared to the model trained on BUPT$_{sub}$ $\cup$ GC$_{bal}$.

Training with data augmentation (Tab. \ref{tab:results-fairness-AUG}) had a generally negative impact on models trained solely on authentic data (Tab. \ref{tab:results-fairness-AUG}, first and third groups), worsening both average verification accuracy and fairness metrics on both the employed authentic datasets. This trend was consistent across the study, with data augmentation resulting in a substantial deterioration of fairness metrics for all models, except for the STD in models trained on DC$_{bal}$, IDF$_{bal}$, and BUPT$_{sub}$ $\cup$ DC$_{bal}$. Interestingly, training with data augmentation led to gains in accuracy across all models trained on combined or synthetic datasets (Tab. \ref{tab:results-fairness-AUG}, second and fourth groups), exception made for BUPT$_{sub}$ $\cup$ DC$_{bal}$.

\hlbox{RQ3}{Training on balanced datasets slightly improved fairness metrics but often resulted in reduced accuracy, particularly when using authentic-only data. However, synthetic data, especially when combined with balanced authentic datasets, shows promising outcomes in both accuracy and fairness. Data augmentation typically introduces trade-offs, as it tends to negatively impact fairness, even though it may provide a modest increase in overall verification accuracy.}

\section{Conclusion and Future Work}  \label{sec:conclusions}
In this paper, we explored the impact of using combined authentic and synthetic datasets on both verification accuracy and fairness of FR models by balancing their demographic representation. Our results revealed that training an FR model with an equal amount of demographically balanced authentic and synthetic data can help reduce the accuracy gap. For example, training on BUPT$_{sub}$ $\cup$ DC$_{bal}$ and BUPT$_{sub}$ $\cup$ IDF$_{bal}$ achieved performances comparable to FR models trained solely on the authentic CASIA-WebFace dataset, with the model trained on BUPT$_{sub}$ $\cup$ DC$_{bal}$ showing a difference of only 3.53\%.
Our study also suggests that training an FR model on a mix of synthetic and authentic demographically balanced datasets can result in a fairer model with lower standard deviation and skewed error ratio. For instance, the model trained on BUPT$_{sub}$ $\cup$ IDF$_{bal}$ achieved an STD of 2.33 and a SER of 1.07, the lowest overall in both metrics. However, the analyses also produced some ambiguous results, where FR models trained on unbalanced datasets achieved better fairness outcomes than those trained on balanced ones.
Finally, we found that while data augmentation typically increases average verification accuracy, it also leads to a rise in standard deviation and skewed error, thereby worsening models' fairness.

Building upon the findings and limitations of this work, our future efforts will focus on exploring the performance of different combinations using a broader range of architectures, such as ResNet-34 and ResNet-100, as well as various loss functions, including ArcFace and AdaFace. Additionally, we plan to incorporate more advanced data augmentation techniques, refined sampling strategies, domain generalization methods, and active learning and/or knowledge distillation techniques to further enhance the accuracy and fairness of the FR models through an optimized combination of both authentic and synthetic data.

\bibliographystyle{splncs04}
\bibliography{main}
\end{document}